\documentclass{article}

\usepackage{PRIMEarxiv}

\usepackage[utf8]{inputenc} 
\usepackage[T1]{fontenc}    
\usepackage{hyperref}       
\usepackage{url}            
\usepackage{booktabs}       
\usepackage{amsfonts}       
\usepackage{nicefrac}       
\usepackage{microtype}      
\usepackage{lipsum}
\usepackage{siunitx}
\usepackage{fancyhdr}       
\usepackage{graphicx}       
\usepackage{algorithm}
\usepackage{algpseudocode}
\graphicspath{{media/}}     

\title{Greedy Clustering-Based Algorithm for Improving Multi-point Robotic Manipulation Sequencing
}

\author{
  Gavin Strunk \\
  \texttt{gavin.strunk@gmail.com} \\
  }

\begin{document}
\maketitle

\begin{abstract}
The problem of optimizing a sequence of tasks for a robot, also known as multi-point manufacturing, is a well-studied problem.  Many of these solutions use a variant of the Traveling Salesman Problem (TSP) and seek to find the minimum distance or time solution.  Optimal solution methods struggle to run in real-time and scale for larger problems. In online planning applications where the tasks being executed are fast, the computational time to optimize the ordering can dominate the total execution time.  The optimal solution in this application is defined as the computational time for planning plus the execution time.  Therefore, the algorithm presented here balances the quality of the solution with the total execution time by finding a locally optimal sequence.  The algorithm is comprised of waypoint generation, spatial clustering, and waypoint optimization.  Significant improvements in time reduction were seen and validated against a base case algorithm in simulation and on a real UR5 robot. 
\end{abstract}

\section{Introduction}
There are many manufacturing processes that can improve in speed and reliability with the help of modern day robotics.  These applications commonly require that a robot visits many points in space and performs an action, such as welding, spraying, or drilling.  These problems, known as multi-point, have been studied extensively to find optimal formulations and solution methods\cite{Alatartsev2015}\cite{Zhang2016}.  Applications that require online planning and execute on the order of minutes pose additional difficulties because of the computation requirements of many optimal approaches.  These applications need to consider the planning time plus the execution time to ensure the overall system performance is optimized.  This becomes an interesting tradeoff between quality of the solution (reduces execution time) and computational time to find the solution.  

The application presented in this paper also becomes more complicated because it is assumed that the points are not all reachable by the 6 DOF robot.  Therefore, a turntable is used to present points in a space that the robot can reach.  Therefore, this problem cannot be described as a classical Traveling Salesman Problem (TSP).  There are variations of the TSP that can describe this problem and will be discussed in detail in the related work.  One method to model the problem is to consider the turntable to be an additional joint of the robot.  However, the turntable constraints make this a complex problem to solve in real time because it has a velocity that is significantly slower than the other joint velocities of the robot, and is only able to rotate in one direction.  Formulating the problem in this manner would result in a highly nonlinear problem and constraints, which has not shown to be readily solved within the time constraints required by this application.  The final assumption of this work is the sequence and motion planning are handled separately, not combined as done in \cite{Kovacs2016}.  This is done to ensure that obstacles are still avoided for safe execution and there are a variety of mature motion planning options, such as Probabilistic Roadmap (PRM) and Rapidly-exploring Random Tree (RRT)\cite{Elbanhawi2014}.

The contribution of this work was to develop an algorithm that can find a locally optimal sequencing that improves the overall execution time of a multi-point problem with a 6 DOF robotic manipulator and turntable, including the planning time.  The remainder of the paper is organized as follows.  Section II describes related work in terms of various problem formulations and solution methods.  Section III details the algorithm.  Section IV provides the results from both simulation and real robot experiments, and lastly section V concludes.

\section{Related Work}
There has been a significant amount of research performed to solve the task sequencing problem.  A majority of the novelty comes from the subtlety in the problem formulation and the method used to solve the problem.  A common problem formulation is the classical TSP as done in \cite{Zhang2016}\cite{Dubowsky1989}\cite{Zacharia2005}\cite{Chen2017}\cite{Xidias2010}\cite{Zacharia2013}.  The main commonality for this formulation is neglecting the inverse kinematics (IK) of the robot.  A variety of methods are used to accomplish this, such as only considering 3 DOF \cite{Zhang2016}.  The important points for the TSP formulation are the assumption that the robot has to return to the initial state and only visit each point once, multiple inverse kinematic (IK) solutions to reach the same point are not considered, and obstacles in the environment are ignored as well.  Solving this formulation has been commonly done with a genetic algorithm (GA) \cite{Zhang2016}\cite{Xidias2010}\cite{Zacharia2005}\cite{Zacharia2013}.  These methods have shown to scale well with the problem size compared to other methods used, but they still require a significant computational time.  Other solution approaches have been particle swarm optimization \cite{Chen2017}, ant colony optimization \cite{Saenphon2014}, and branch and bound algorithms \cite{Dubowsky1989}\cite{Ferone2016}.

Another popular formulation is referred to as Robotic Task Sequencing Problem (RTSP) \cite{Suarez-Ruiz2017} or Generalized Traveling Salesman Problem \cite{Carlson2014}.  The RTSP extends the classical TSP by making the assumption that each of the locations may be reached in more than one configuration.  The GTSP accomplishes a similar mechanism by thinking of the IK solutions as a bin and the robot is required to visit one solution per bin.  The solution methods to these problems are similar to the TSP such as branch and bound or a task space method by \cite{Suarez-Ruiz2017}.

The next approach is a Traveling Salesman Problem with Neighborhoods (TSPN) \cite{Kovacs2016}\cite{Alatartsev2013} or similarly the Touring Polygon Problem (TPP) \cite{Alatartsev2014}.  The main difference between these and the GTSP is the points are still grouped into neighborhoods, but the robot is only required to visit the boundary of each group rather than a point in the neighborhood.  The TPP simply does not require the robot to return to the initial location.  The Rubber-band algorithm has been successfully applied to these problems as well as a greedy approach used by \cite{Kovacs2016}.

Other approaches, unrelated to a TSP, to the problem include phase plane \cite{Bobrow1985} and constraint multi-objective optimization \cite{Kolakowska2014}\cite{HepingChen} formulations.  The phase plane approach involves minimizing the time of the robot movement by trying to maximize the velocity.  A constraint is constructed based on the dynamics that defines a boundary for the maximum tolerable velocity.  Assuming the robot's velocity is zero at the start and end point they follow a path of maximum acceleration until they reach the boundary.  If the boundary is crossed then a path from the end is projected back using the maximum deceleration, and the point where they cross is the switching point of the phase.

The constraint multi-objective optimization formulation is used when you try to minimize the travel time of the robot, as well as, an application specific objective like spray coverage.  This approach works well if the solution to minimize the robot's time is only valid if another constraint is satisfied.  Solutions to these problems have taken the form of converting the problem to an unconstrained single object with a method like weighted sum \cite{HepingChen}.

In this paper we will formulate the problem as a modified Shortest Sequence Problem (SSP)\cite{Alatartsev2015}, also known as Shortest Path Problem (SPP)\cite{Festa2013}\cite{Saha}, which is similar to a TSP except the robot is not required to return to the initial location.  The deviation from the classic SSP is the assumption that the points are only reachable using the turntable and the 6 DOF robot, which is not modeled as a redundant joint of the robot.  It is also important to remember the goal is to minimize the total time (planning plus execution), so the computational cost of the algorithm represents only the planning time.  

\section{Algorithm}
\begin{figure}[ht]
	\begin{center}
	\includegraphics[width=3in]{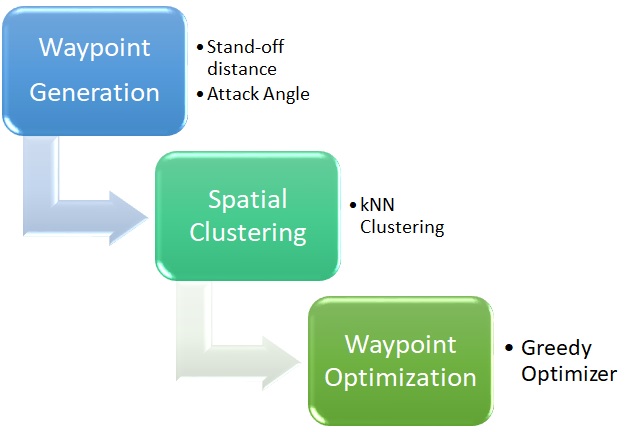}
	\caption{Algorithm Block Diagram}
	\label{fig:algorithm}
	\end{center}
\end{figure}

In this section we will describe the algorithm used to solve the SSP.  A block diagram of the algorithm is shown in Figure \ref{fig:algorithm}, which is comprised of three portions: a waypoint generator, spatial clustering, and a waypoint optimizer.  These will be discussed in more detail, but it is worth noting that these components can be interchanged depending on the application.  This provides an algorithm that is flexible to a variety of applications fitting the basic assumptions regardless of the application itself.  

\subsection{Waypoint Generation}
\begin{figure}[ht]
	\begin{center}
	\includegraphics[width=2in]{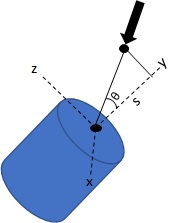}
	\caption{Algorithm Block Diagram}
	\label{fig:waypointgen}
	\end{center}
\end{figure}

The first step in the algorithm is to generate the waypoints the robot needs to visit.  The waypoint generator does this by taking in locations of objects and outputting the desired position and orientation SE(3) of the end effector.  Generating the waypoints could be done in a variety of ways and it is typically application specific.  It is common in spraying, drilling, and welding tasks that the waypoints can be generalized by a stand-off distance and an attack angle that are referenced to a coordinate frame located on the piece being worked on as shown in Figure \ref{fig:waypointgen}.  For this application, coordinate frames are located in the center of each hole and the y axis is oriented to pass through the centerline of the hole.  The rest of the axes are defined using the right-hand rule.  Therefore, the stand-off distance $s$ is defined as a translation along the y axis and the positive attack angle $\theta$ is a counter-clockwise rotation about the x axis.

In addition to the SE(3) vector, an angle relative to the center of rotation of the turntable is also associated with each waypoint.  This angle is important because it relates the points location in space relative to the current angle of the turntable.  This results in a group of waypoint and angle pairs that are randomly ordered, and are not guaranteed to be within reach of the robot.  

\subsection{Spatial Clustering}
\begin{figure}[ht]
	\begin{center}
	\includegraphics[width=3in]{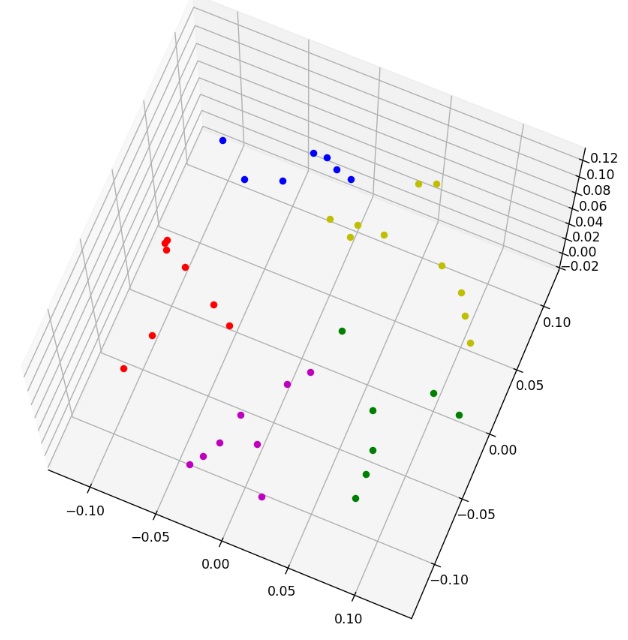}
	\caption{k-NN Clustering on 3D points}
	\label{fig:clustering}
	\end{center}
\end{figure}

After the waypoints and angles are generated, the algorithm needs a way to get the waypoints to a location that the robot can reach.  One way to approach this task would be to rotate the turntable to the waypoint's associated angle, but the turntable is significantly slower than the robot so this is not an efficient choice.  Therefore, making the waypoints reachable was accomplished using spatial clustering based on the heuristic knowledge of an angular bound that the robot can reliably reach.  This was chosen to be \ang{72} which sections the object on the turntable into five sections.  This was used as the number of clusters to calculate for the k-Nearest Neighbors (k-NN) clustering algorithm\cite{Chen2015}.  The results of the clustering can be seen in Figure \ref{fig:clustering}.

The output of this algorithm is a non-deterministic group of clustered waypoints in space.  To minimize the turntable movement, the average angle relative to the turntable coordinate axes is calculated.  This was done for each cluster using the mean of circular quantities in Equation \ref{eq:circmean}. 

\begin{equation}
	\alpha = atan2(\sum x_i, \sum y_i)
	\label{eq:circmean}
\end{equation}

The turntable uses this angle to center the cluster to the operating boundary of the robot.  Finally, the clusters are sorted by ascending average angle so the turntable only must make one rotation to present every point to the robot.  Now we have a group of clusters that ensure the points are reachable by the robot, but they are still randomly sorted within each of the clusters.     

\subsection{Waypoint Optimization}
The next step is to improve the points sequence to minimize the distance within each cluster.  Therefore, the clusters are treated as sub problems solving for a locally optimal solution within each cluster.  This was done with a greedy algorithm described in Algorithm \ref{alg:greedy}. The first step is to calculate the distance matrix between the points\cite{Wang2018}, which creates a fully connected graph.  It is assumed that the robot can travel from any point to any point, except itself, within a cluster.  The algorithm starts by assuming the first point in the sequence is the starting point.  It then finds the minimum distance to travel and chooses that point.  To ensure points are only visited once, edges leading to that point are set to a high value to ensure it will not travel any path that leads back to that point.  This process continues until all points in the cluster are visited.  

\begin{algorithm}
	\caption{Greedy Waypoint Optimizer}
	\label{alg:greedy}
	\begin{algorithmic}[1]
	\State $Calculate M_d$
	\State $Assign$ main diagonal to $MAXVAL$
	\State $Assign$ first column to $MAXVAL$
	\While {$x < length(M_d)$}
		\State $i \gets argmin(M_d(:,i))$
		\State $M_d[i,:] = MAXVAL$
		\State $append(i)$
	\EndWhile
	\State $Sort(points)$ 
	\end{algorithmic}
\end{algorithm}

\section{Experimental Results}
\begin{figure}[ht]
	\begin{center}
	\includegraphics[width=3in]{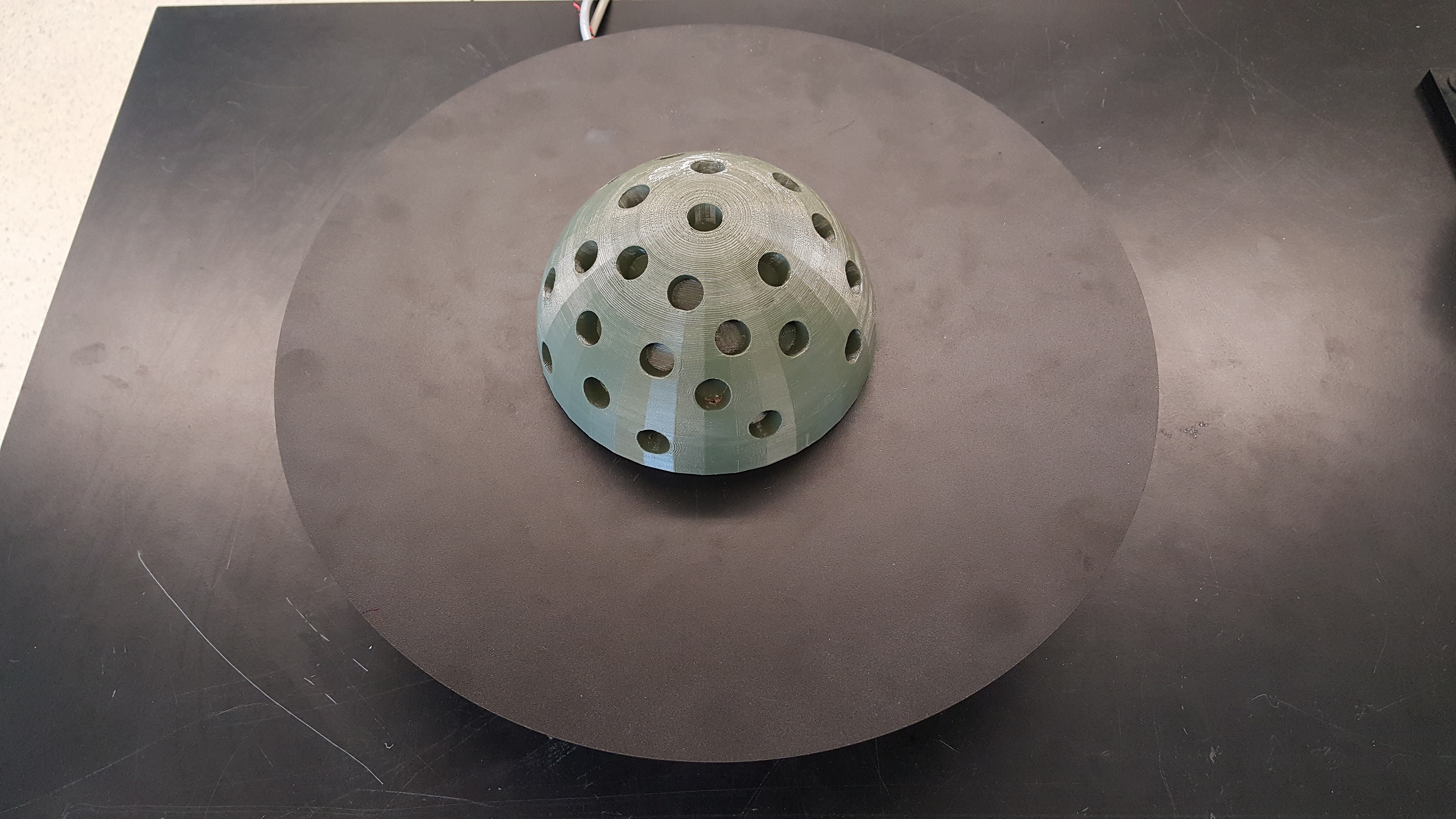}
	\caption{Image of hemispheric object used for experiments.}
	\label{fig:object}
	\end{center}
\end{figure}

A combination of simulation and real robot experiments were run to validate the performance of the algorithm.  The object used for the test was a hemisphere with holes drilled at known locations and orientations shown in Figure \ref{fig:object}.  This created a use case analogous to a 3D peg-in-hole setup, but rather than inserting something into the hole it is assumed that the hole is sprayed.  The part had 40 holes drilled, which was chosen to be sufficiently large so computational cost restrictions would be apparent.  It was also assumed that the part was centered on the turntable and does not move relative to the turntable throughout the process.  PRM* was the algorithm chosen for the motion planner primarily because it empirically proved to reliably find smooth efficient paths.  It was bounded at 5 seconds of search time, which was enough to find efficient paths.  The implementation was done using ROS and the motion planner was taken from MOVEIT.  The simulation and physical experiments were both performed on a UR5 robot.

The benchmark criteria used to compare the results were the planning time, total execution time, SSP distance, angular distance, and travel distance.  The planning time is used to measure the computational cost of using the algorithm.  The execution time is the planning plus execution time, and is the quantity that we are ultimately trying to minimize.  The SSP distance is defined as a linear distance calculation between the waypoints not accounting for objects in the path, inverse kinematics, or the motion planner.  This measures the pure improvement to the sequencing using the algorithm.  The angular distance was measured with an execution monitor that accumulated the total angular travel of all the joints.  The travel distance is the accumulation of the 3D positional travel of the end effector.  The lower bound of the travel distance is the SSP distance as this would represent no objects in the path and pure linear paths.  This does not however directly relate to the angular distance.  The angular and travel distance are measurements of the motion planner's effort required to reach the points in the sequence provided.  These quantities were both used to measure the effect of the sequence on the motion planner's performance, which directly affects the execution time. 

\subsection{Simulated UR5 Arm}

\begin{table*}[!htb]
	\begin{center}
	\caption{Simulation results}
	\label{tab:simresults}
	\begin{tabular}{|l|c|c|c|c|c|}
		\hline
		Algorithm & Planning Time (s) & Execution Time (mm:ss) & SSP Distance (m) & Angular Distance (rad) & Travel Distance (m) \\
		\hline
		Angle 1 & 0.006 & 8:12 & 2.527 & 327.58 & 39.45 \\
		\hline
		Angle 2 & 0.006 & 7:46 & 2.527 & 294.91 & 49.11 \\
		\hline
		Angle 3 & 0.007 & 7:28 & 2.527 & 272.29 & 36.56 \\
		\hline
		Position 1 & 0.029 & 6:42 & 2.454 & 175.08 & 23.24 \\
		\hline
		Position 2 & 0.028 & 6:24 & 2.483 & 160.94 & 25.38 \\
		\hline
		Position 3 & 0.033 & 6:37 & 2.446 & 171.76 & 18.07 \\
		\hline
		Greedy 1 & 0.034 & 6:25 & 1.893 & 182.83 & 21.87 \\
		\hline
		Greedy 2 & 0.030 & 5:53 & 1.728 & 102.48 & 7.65 \\
		\hline
		Greedy 3 & 0.028 & 6:11 & 1.784 & 141.41 & 16.70 \\
		\hline
		Greedy RRT & 0.033 & 3:30 & 2.473 & 154.40 & 19.39 \\ 
		\hline
	\end{tabular}
	\end{center}
\end{table*}

Each algorithm was run three times to get an average value for the benchmark criteria.  The summary of the results is shown in Table \ref{tab:simresults}.  The algorithms tested were angle clustering, position clustering, and the full algorithm presented with clustering and the greedy optimizer.  The angle clustering simply grouped the points into five groups based on the angle relative to the turntable.  They were split into five groups because this was also chosen for the number of clusters, and we did not want to add significant time to the execution due to the turntable moving for each point.  Therefore, this is taken as the base case if no ordering is performed to the points beyond ensuring they are reachable.  The next algorithm is position clustering only.  This was benchmarked to measure the performance gain of the clustering and the greedy algorithm independently.  Finally, the full algorithm presented was run with PRM* and RRT as the motion planner.

The angle algorithm had an average planning time of 0.006 s and resulted in an average execution time of 7 min 49 s.  It can also be seen that the SSP distance is the same for each run.  This was a validation check because the grouping of the points was deterministic so the linear distance was expected to be the same.    It is also interesting to note the order of magnitude change between the theoretical minimum distance and the actual distance traveled.  This would suggest the unordered sequence of points is not only sub-optimal, but also difficult for the motion planner to create efficient trajectories.

The position clustering showed an improvement of 16\% resulting in an average execution of 6 min 34 s, while only requiring 0.03 s to compute.  Improvements were seen in the reduction of all the benchmark criteria compared to the base case.  The SSP distance was no longer constant because the k-NN clustering is non-deterministic.  

Next, the full algorithm was tested and resulted in an average execution time of 6 min and 10 s or 21.1\% improvement over the base case, while requiring 0.031 seconds to compute.  It is expected that the computational increase is negligible because the algorithm runs in $O(N)$.  It was surprising to note that only a marginal improvement was observed in the execution time, while the other benchmark criteria were reduced more significantly.  After further investigation, it was realized that the bottleneck was the time required for the PRM* to compute the point-to-point trajectory.  It was using around 5 s per point resulting in 200 s of motion planning overhead.  The test was run again with the RRT motion planner and a \textbf{55.2\%} improvement (3 min and 30 s) over the base case was observed.  The RRT motion planner was not used originally because it could not reliably plan every point-to-point trajectory, but because the algorithm reduced the distance between the points the RRT was now able to plan every trajectory.  The next step was to validate the findings on a real robot. 

\subsection{Real UR5 Arm}
\begin{figure}[th]
	\begin{center}
	\includegraphics[height=2in, angle=270]{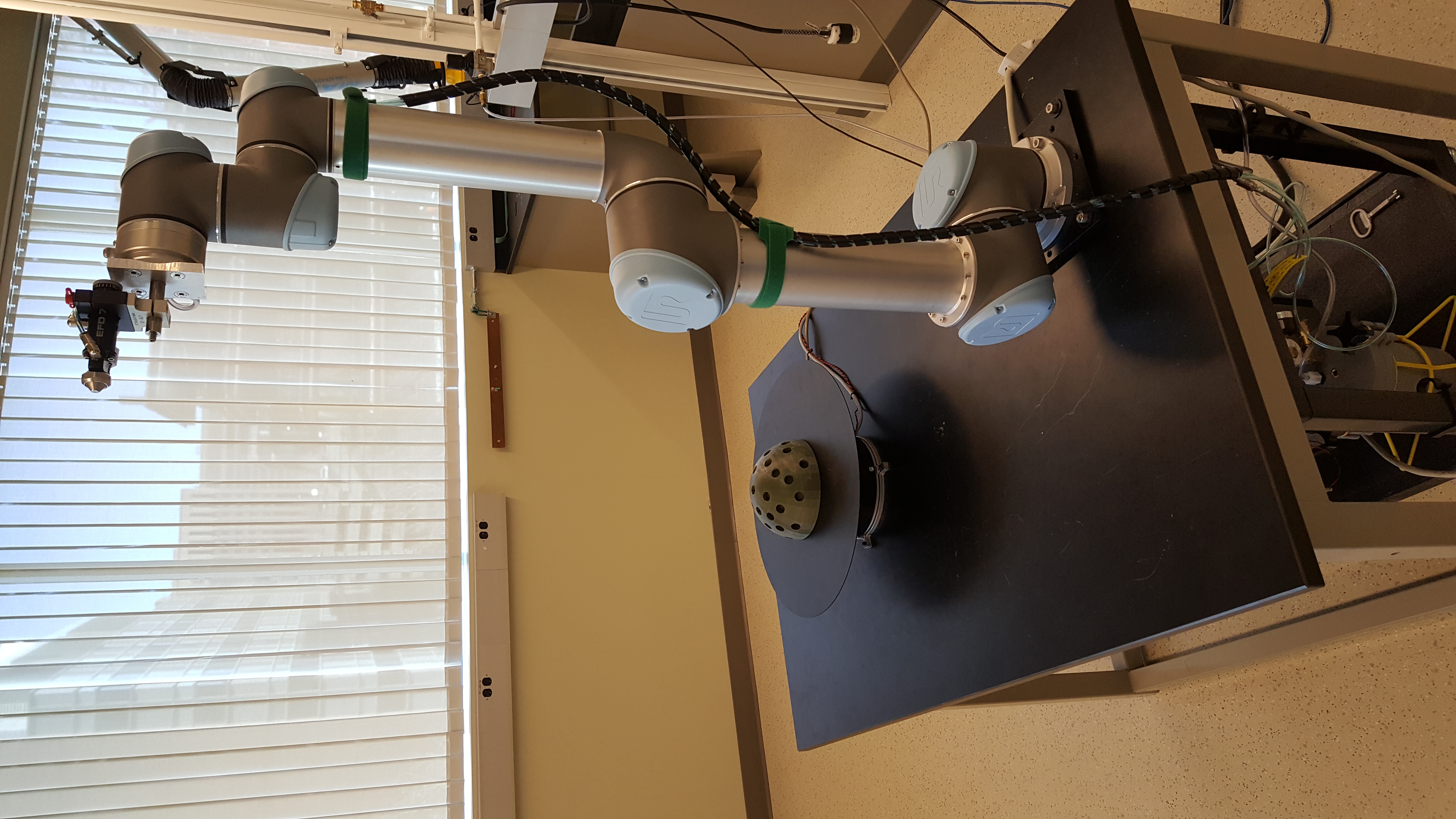}
	\caption{UR5 work cell used for experiments.}
	\label{fig:cell}
	\end{center}
\end{figure}

To validate the algorithm's results, a real physical setup was constructed that matched the simulation environment and is shown in Figure \ref{fig:cell}.  

The results from the real robot matched well with the results from the simulation.  The test configurations and the resulting execution times are shown in Table \ref{tab:realresults}.  The first test was using the base case mentioned previously, and resulted in a time of 8 min 17 s compared to 7 min 49 s average from the simulation.  The algorithm with PRM* as the motion planner showed similar improvements of 19.9\% compared to 21.1\% for the simulation.  Finally, the algorithm running with RRT as the motion planner had the same efficiency gains (58.8\% compared to 55.2\%) and could plan trajectories to the points.  A base case test was not included using RRT because it was found that it was not able to successfully plan trajectories for every point. 

\begin{table}[htbp]
	\begin{center}
	\caption{UR5 Results}
	\label{tab:realresults}
	\begin{tabular}{|l|c|}
		\hline
		Algorithm & Execution Time (mm:ss) \\
		\hline
		No Optimization w/ PRM* & 8:17 \\
		\hline
		Optimization w/ PRM* & 6:38 \\
		\hline
		Optimization w/ RRT & 3:25 \\
		\hline
	\end{tabular}
	\end{center}
\end{table}

\section{Conclusion}
An algorithm was developed to address the sequencing problem for a real-time application that also had to account for points being out of the robot's reach.  It was shown to significantly improve the execution time compared to a simple angular grouping in simulation and was further validated on a physical setup.  Another benefit of the algorithm was making it easier for the motion planner to find trajectories, which allowed the use of a faster trajectory planner resulting in another significant reduction in execution time.  The algorithm is non-deterministic, which results in some variance in the performance gains.  

One way to try reducing the variance would be to run parallel clustering steps and choose the sequence that results in the minimum SSP distance.  This would likely improve the variance, but would not guarantee lower execution time because SSP distance and execution time are not directly related.  It is observed from the results that the angular distance relates directly to the execution time, but this would be impractical to compute online because it depends on the inverse kinematics solution and the motion planner trajectory.  Overall, the algorithm provided a computationally inexpensive solution that significantly improve the execution time.

\bibliographystyle{unsrt}
\bibliography{References}

\end{document}